\documentclass{llncs}
\usepackage{amsmath,graphicx}
\usepackage{amsmath,amssymb} 
\usepackage{epsfig}
\usepackage{multirow}
\usepackage{llncsdoc}
\usepackage{etoolbox}

\begin{document}

\mainmatter

\title{AMD Severity Prediction And Explainability Using Image Registration And Deep Embedded Clustering}

\titlerunning{}

\author{Dwarikanath Mahapatra and Hidemasa Muta }


\institute{IBM Research Australia \\
\email{[dwarim,hidem]@au1.ibm.com.}}



\maketitle

\begin{abstract}
We propose a method to predict severity of age related macular degeneration (AMD) from input optical coherence tomography (OCT) images. Although there is no standard clinical severity scale for AMD, we leverage  deep learning (DL) based image registration and clustering methods to identify diseased cases and predict their severity. Experiments demonstrate our approach's disease classification performance matches state of the art methods. The predicted disease severity performs well on previously unseen data. Registration output provides better explainability than class activation maps regarding label and severity decisions.    
\end{abstract}


\section{Introduction}
\label{sec:intro}

Most approaches to deep learning (DL) based medical image classification output a binary decision about presence or absence of a disease without explicitly justifying decisions. Moreover, disease severity prediction in an unsupervised approach is  not clearly defined unless the labels provide such information, as in diabetic retinopathy \cite{Kaggle}.
%
Diseases such as age related macular degeneration (AMD) do not have a standard clinical severity scale and it is left to the observer's expertise to assess severity. While class activation maps (CAMs) \cite{CAM} highlight image regions that have high response to the trained classifier they do not provide measurable parameters to explain the decision. Explainability of classifier decisions is an essential requirement of modern diagnosis systems.


In this paper we propose a convolutional neural network (CNN) based optical coherence tomography (OCT) image registration method that: 1) predicts the disease class of a given image (e.g., normal, diabetic macular edema (DME) or dry AMD); 2) uses registration output to grade disease severity on a normalized scale of $[1,10]$ where $1$ indicates normal and $10$ indicates confirmed disease, and 3) provides explainability by outputting measurable parameters.

Previous approaches to DL based image registration include regressors \cite{Vos_DIR,RegNet,Mahapatra_GAN_CVIU2019,Mahapatra_ISR_CMIG2019,Mahapatra_LME_PR2017,Zilly_CMIG_2016,Mahapatra_SSLAL_CD_CMPB,Mahapatra_SSLAL_Pro_JMI,Mahapatra_LME_CVIU,LiTMI_2015,MahapatraJDI_Cardiac_FSL,Mahapatra_JSTSP2014} and generative adversarial networks (GANs) \cite{Mahapatra_MLMI18,MahapatraTIP_RF2014,MahapatraTBME_Pro2014,MahapatraTMI_CD2013,MahapatraJDICD2013,MahapatraJDIMutCont2013}. 
\cite{BalaCVPR18,MahapatraJDIGCSP2013,MahapatraJDIJSGR2013,MahapatraJDISkull2012,MahapatraTIP2012,MahapatraTBME2011} learn a parameterized registration function from training data without the need for simulated deformations in \cite{RegNet,MahapatraEURASIP2010,Mahapatra_ISBI19,MahapatraAL_MICCAI18,Mahapatra_MLMI18,Mahapatra_MICCAI17}. 
Although there is considerable research in the field of interpretable machine learning their application to medical image analysis  problems is limited \cite{BrainImimic18,PathImimic18,Mahapatra_MLMI16,Mahapatra_EMBC16,Mahapatra_MLMI15_Optic,Mahapatra_MLMI15_Prostate,Mahapatra_OMIA15,MahapatraISBI15_Optic}. The CAMs of \cite{CAM} serve as visualization aids rather than showing quantitative parameters. We propose a novel approach to overcome the limitations of CAM, by providing quanitative measures and their visualization for disease diagnosis based on image registration. Image registration makes the approach fast and enables projection of registration parameters to a linear scale for comparison against normal and diseased cases. It also provides localized and accurate quantitative output compared to CAMs.
 Our paper makes the following contributions: 1) a novel approach for AMD severity estimation using registration parameters and clustering; and 2) mapping registration output to a classification decision and output quantitative values explaining classification decision.

%

\section{Method}
\label{sec:method}

Our proposed method consists of: 1) atlas construction for different classes; 2) End to end training of a neural network to estimate registration parameters and assign severity labels; 3) Assign a test volume to a disease severity scale, output its registration parameters and provide quantitatively interpretable information.

\subsection{Atlas Construction Using Groupwise Registration}

All normal volumes are coarsely aligned using their point cloud cluster and the  iterated closest point (ICP) algorithm. Groupwise registration using ITK \cite{ITK} on all volumes gives the atlas image $A_N$. Each normal image is registered to $A_N$ using B-splines. The registration parameters are displacements of grid nodes. They are easier to store and predict than a dense $3$D deformation field and can be used to generate the $3$D deformation field. The above steps are used to obtain atlases for AMD ($A_{AMD}$) and DME ($A_{DME}$). 
%





\subsection{Deep Embedded Clustering Network}

%

Deep embedded clustering \cite{DEC,MahapatraISBI15_JSGR,MahapatraISBI15_CD,KuangAMM14,Mahapatra_ABD2014,MahapatraISBI_CD2014,MahapatraMICCAI_CD2013} is an unsupervised clustering approach and gives superior results than traditional clustering algorithms.  
To cluster n points ${x_i \in X}^{n}_{i=1}$ into k clusters, each represented by a centroid $\mu_j , j = 1,\cdots,k$, DEC first transforms the data with a nonlinear mapping $f_{\theta}:X \mapsto Z$, where $\theta$ are learnable parameters and $Z$ is the latent feature space with lower dimensionality than $X$. Similarity between embedded point $z_i$ and cluster centroid $\mu_j$ is given by the Student's t-distribution as 
\begin{equation}
q_{ij}=\frac{\left(1+\left\|z_i-\mu_j \right\|^{2}/\alpha \right)^{-\frac{\alpha+1}{2}}} {\sum_{j'} \left(1+\left\|z_i-\mu_{j'} \right\|^{2}/\alpha \right)^{-\frac{\alpha+1}{2}}},
\end{equation}
where $\alpha=1$ for all experiments. 
DEC simultaneously learns $k$ cluster centers in feature space $Z$ and the parameters $\theta$. It involves: (1) parameter initialization with a deep autoencoder \cite{Vincent2010} and (2) iterative parameter optimization by computing an auxiliary target distribution and minimizing the Kullback–Leibler (KL) divergence. For further details we refer the reader to \cite{DEC}

\subsection{Estimation of Registration parameters}

\begin{figure}[h]
\centering
\includegraphics[height=6.0cm, width=11cm]{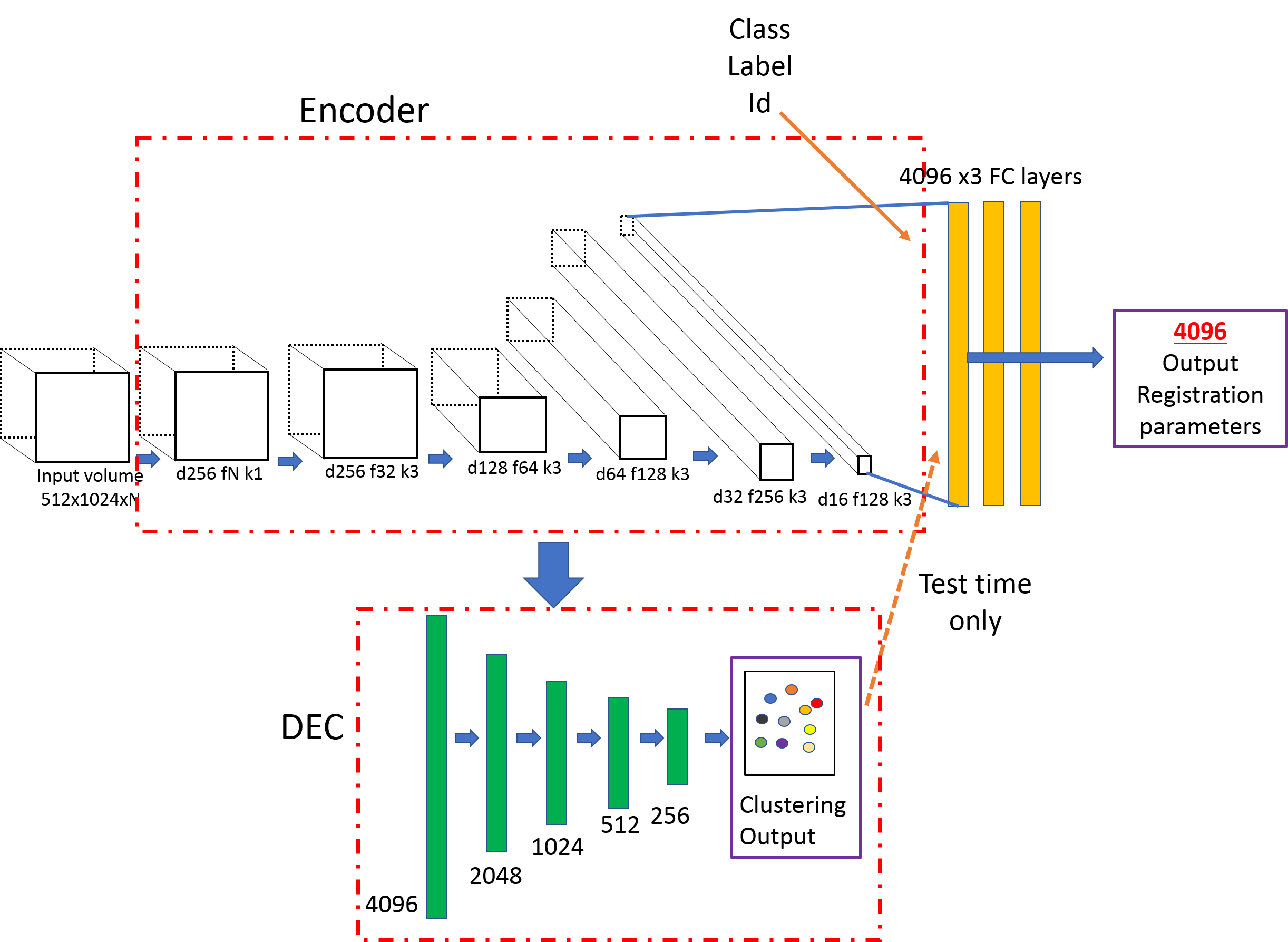}
  \caption{Architecture of our proposed network for AMD classification and severity estimation. A regression network for image registration and deep embedded clustering network are combined to achieve our objectives.}
\label{fig:workflow}
\end{figure}

Conventional registration methods output a deformation field from an input image pair while we jointly estimate the grid displacements and severity label using end to end training. 
Figure~\ref{fig:workflow} depicts our workflow. An input volume of dimension $512\times1024\times N$, $N$ is number of slices, is converted to a stack of $N$ convolution feature maps by downsampling to $256\times512\times N$ and employing $1\times1$ convolution. The output is shown in Figure~\ref{fig:workflow} as $\text{d256 fN k1}$, which indicates output maps of dimension ($d$) $256\times512$, $N$ feature maps ($f$) and kernel dimension ($k$) of $1\times1$. The next convolution layer uses $3\times3$ kernels and outputs $f=32$ feature maps. This is followed by a max pooling step that reduces the map dimensions to $128\times128$ and the next convolution layer outputs $64$ feature maps using $3\times3$ kernels. After three further max pooling and convolution layers, the output of the ``Encoder'' stage are $128$ feature maps of dimension $16\times16$.

The Encoder output is used in two ways. The first branch is the input to the Deep Embedded Clustering (DEC) network (green boxes depicting fully connected layers) that outputs a cluster label indicating severity score. The second branch from the Encoder is connected, along with the input volume's disease label, to a fully connected (FC) layer (orange boxes) having $4096$ neurons. It is followed by two more FC layers of $4096$ neurons each and the final output is the set of registration parameters. The ``Class Label id'' (disease label of input volume) and the Encoder output are combined using a global pooling step. 
The motivation behind combining the two is as follows: We are interested to register, for example, a normal volume to the normal atlas. The ground truth registration parameters of a normal volume correspond to those obtained when registering the input volume to the normal atlas, and we want the regression network to predict these parameters. Feeding the input volume's actual disease label guides the regression network to register the image to the corresponding atlas.



\subsection{Training Stage Implementation}

The entire dataset is divided into training ($70\%$), validation ($10\%$) and test ($20\%$) folds for each class. 
The DEC parameter initialization closely follows the steps outlined in \cite{DEC}. The regression network is trained using the input images, their labels and the corresponding registration parameters. We augment the datasets $150$ times by rotation and flipping and obtain their registration parameters with the corresponding atlas. In the first phase of training only the regression network is trained using mean squared error (MSE) loss for $50$ epochs to get an initial set of weights. 
Subsequently, the DEC is trained using the output of the Encoder network. After training is complete we cluster the different volumes and observe that $97.8\%$ of the normal patients are assigned to clusters $1,2$ and $3$. $97.5\%$ of Diabetic macular Edema (DME) cases are assigned to clusters $4,5,6$ and $7$, while $97.2\%$ of AMD cases are assigned to clusters $8,9$ and $10$. Thus the following mapping between image labels and cluster labels are obtained $\text{Normal} \in \{1,2,3\}, \text{DME} \in \{4,5,6,7\}, ~\text{and~ AMD} \in \{8,9,10\}$. 

\subsection{Predicting Severity of test image}

When a test image comes in we first use the trained DEC to predict the cluster label, which apart from providing disease severity on a scale of $[1,10]$ also gives the image's disease class. 
The disease label is then used to predict the image's registration parameters to the corresponding atlas. 
Depending upon the desired level of granularity of disease severity the number of clusters can be varied to identify different cohorts that exhibit specific traits.

%

\section{Experimental Results}
\label{sec:expts}


We demonstrate the effectiveness of our algorithm on a public dataset \cite{AMDData,MahapatraProISBI13,MahapatraRVISBI13,MahapatraWssISBI13,MahapatraCDFssISBI13,MahapatraCDSPIE13,MahapatraABD12} consisting of OCT volumes from 50 normal,  48  dry  AMD,  and  50  DME patients. The  axial  resolution of the images is  $3.5$ $\mu$-m with   scan dimension  of $512\times1024$ pixels. The  number  of  B-scans  varies between $19,  25,  31, 61$ per  volume in different  patients.  The  dataset is publicly available at http://www.biosigdata.com. For all registration steps we used a grid size of $16\times16\times16$. The number of predicted grid parameters are $16^{3}=4096$



 
\subsection{Registration Results}

The output registration parameters from our method are used to generate a deformation field using B-splines and compared with outputs of other registration methods. For the purpose of quantitative evaluation we applied simulated deformation fields and use different registration methods to recover the registration field. Validation of accuracy is based on mean absolute distance (MAD) between applied and recovered deformation fields. We also manually annotate retinal layers and compute their $95\%$ Hausdorff Distance ($HD_{95}$) and  Dice Metric (DM) before and after registration.
Our method was implemented with Python and Keras, using SGD and Adam with $\beta_1=0.93$ and batch normalization. Training and test was performed on a NVIDIA Tesla K$40$ GPU with $12$ GB RAM. 

Table~\ref{tab:reg1} compares results of the following methods: 1) $Reg-DEC$: Our proposed method; 2) $Reg_{NoDEC}$: $Reg-DEC$ using only the registration without additional clustering; 3) $VoxelMorph$: The method of \cite{BalaCVPR18,MahapatraMLMI12,MahapatraSTACOM12,VosEMBC,MahapatraGRSPIE12,MahapatraMiccaiIAHBD11,MahapatraMiccai11}; 4) $FlowNet$: - the registration method of \cite{FlowNet,MahapatraMiccai10,MahapatraICIP10,MahapatraICDIP10a,MahapatraICDIP10b,MahapatraMiccai08,MahapatraISBI08}; 5) $DIRNet$: - the method of \cite{Vos_DIR,MahapatraICME08,MahapatraICBME08_Retrieve,MahapatraICBME08_Sal,MahapatraSPIE08,MahapatraICIT06}; 6) $Reg-kMeans$ - replacing DEC with kmeans clustering. Our method outperforms the state of the art DL based registration methods. 


\begin{table}[b]
\begin{tabular}{|c|c|c|c|c|c|c|c|}
\hline
{} & {Bef.} & \multicolumn {6}{|c|}{After Registration}  \\ \cline{3-8} 
{} & {Reg} & {Reg-DEC} & {Reg$_{NoDEC}$}  & {Reg-kMeans} & {DIRNet} & {FlowNet} & {VoxelMorph}  \\ \hline
%
{DM($\%$)} & {78.9} & {89.3}  & {85.9} & {84.8} & {83.5} & {87.6} & {88.0} \\ \hline
{HD$_{95}$(mm)} & {12.9} & {6.9}  & {8.4} & {8.7} & {9.8} & {7.5} & {7.4} \\ \hline
{MAD} & {13.7} & {7.3}  & {8.9} & {10.3} & {9.1} & {8.6} & {7.9} \\ \hline
{Time(s)} & {} & {0.5} & {0.4} & {0.6} & {0.5} & {0.6} & {0.6} \\ \hline
\end{tabular}
\caption{Image registration results from different methods. $Time$ indicates computation time in seconds.}
\label{tab:reg1}
\end{table}

\subsection{Classification Results}

Table~\ref{tab:ClassAMD} summarizes the performance of different methods on the test set for classifying between normal, DME and AMD. Results are also shown for CNN based classification networks such as VGG-16 \cite{VGG}, Resnet \cite{ResNet} and DenseNet \cite{Densenet}, three of the most widely used classification CNNs and the multiscale CNN ensemble of \cite{AMDData} that serves as the baseline for this dataset. Our proposed method outperforms standard CNN architectures, thus proving the efficacy of combining registration with clustering for classification tasks. It also shows $Reg-DEC$'s advantages of lower computing time and fewer training parameters.


\begin{table}[b]
\begin{tabular}{|c|c|c|c|c|c|c|c|}
\hline
{} & {$Reg-DEC$} & {$VGG_{16}$} & {$ResNet_{50}$} & {$DenseNet$} & {DEC} & {kmeans} & {MultCNN \cite{AMDData}} \\  \hline
{Sen} & {93.6} & {91.7} & {92.5} & {92.6} & {89.5} & {85.7} & {92.5} \\ \hline
{Spe} & {94.3} & {92.8} & {93.6} & {93.5} & {90.6} & {86.8} & {93.4} \\ \hline
{AUC} & {96.4} & {94.1} & {95.2} & {95.3} & {91.9} & {87.7} & {95.2} \\ \hline
{Time(h)} & {4.3} & {16.7} & {12.4} & {13.6} & {2.5} & {0.5} & {15.1} \\ \hline
\end{tabular}
\caption{Classification results for AMD, DME and normal on the test set using different networks. Time indicates training time in hours.}
\label{tab:ClassAMD}
\end{table}


\subsection{ Identification of Disease Subgroups And Explainability}

Besides predicting a disease label and severity score, our method provides explainability behind the decision. For a given test image and its predicted registration parameters we calculate its $l_2$ distance from each of the $10$ cluster centers to give us a single value quantifying the sample's similarity with each disease cluster. Let the sample $s$ be assigned to cluster $i \in [1,10]$ and let the corresponding $l_2$ distances of $s$ to each cluster be $d_i$. We calculate a normalized value
\begin{equation}
p_d=\left| \frac{d_i-d_1}{d_{10}-d_i} \right|,
\label{eq:disprob}
\end{equation}
where $p_d$ gives a probability of the test sample reaching the highest severity score. It is also a severity score on a normalized scale of $[0,1]$. Scores from multiple visits help to build a patient severity profile for  analysing different factors behind increase or decrease of severity, as well as the corresponding rate of change. The rate of severity change is an important factor to determine a personalized diagnosis plan. $p_d$ is different from the class probability obtained from a CNN classifier. The classifier probability is its confidence in the decision while $p_d$ gives the probability of transitioning to the most severe stage. 

Tables~\ref{tab:reg1},\ref{tab:ClassAMD}  demonstrate $Reg-DEC$'s superior performance for classification and registration. 
To determine $Reg-DEC$'s effectiveness in predicting disease severity of classes not part of the training data, 
 we train our severity prediction network on normal and AMD images only, leaving out the DME affected images. We keep the same number of clusters (i.e., $10$) as before. Since there are no DME images and number of clusters is unchanged, assignment of images to clusters is different than before. In this case $96.4\%$ of AMD images are assigned to clusters $8,9,10$ which is a drop of $0.8\%$ than the previous assignment while $96.5\%$ of normal samples are assigned to clusters $1,2,3$ which is decrease of $1.3\%$. 
%

We see fewer images in clusters $4,5,6,7$ although the majority of original assignments of normal and AMD cases are unchanged. When we use this trained model on the DME images we find that $96.9\%$ of the images are assigned to  clusters $4,5,6,7$, a decrease of $0.9\%$ from before. 
The above results lead to the following conclusions: 1) $Reg-DEC$'s performance reduces by $0.9\%$ for DME and maximum of $1.3\%$ (for Normal images) when DME images were not part of the training data. This is not a significant drop indicating $Reg-DEC$'s capacity to identify sub-groups that were not part of the training data. 2) Using k-means clustering does not give same performance levels demonstrating that end to end feature learning combined with clustering gives much better results than performing the steps separately. 
$Reg-DEC$ accurately predicts disease severity even though there is no standard severity grading scale. Severity scale  also identifies sub-groups from the population with a specific disease activity. 



\begin{figure}[h]
\begin{tabular}{cccc}
\includegraphics[height=1.3cm, width=3.0cm]{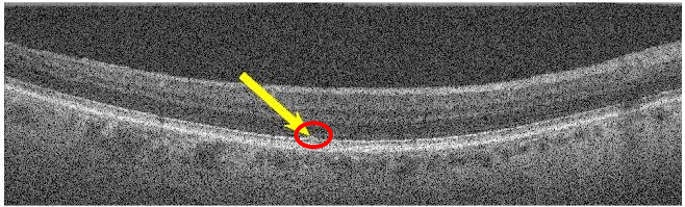} &
\includegraphics[height=1.3cm, width=3.0cm]{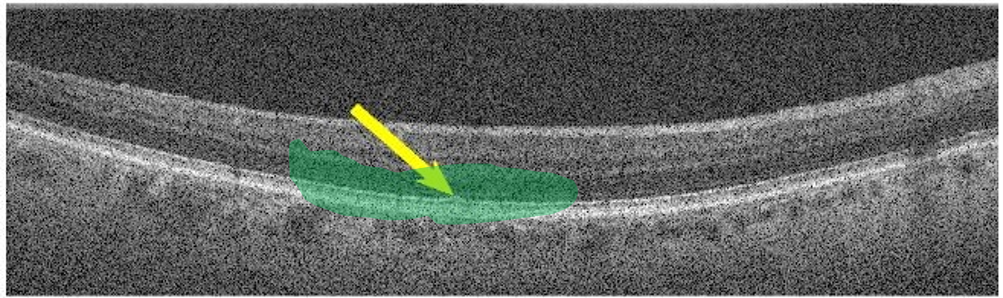} &
\includegraphics[height=1.3cm, width=3.0cm]{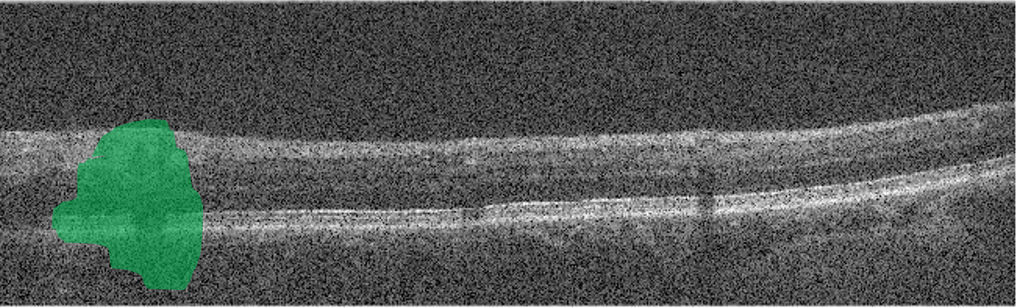} &
\includegraphics[height=1.3cm, width=3.0cm]{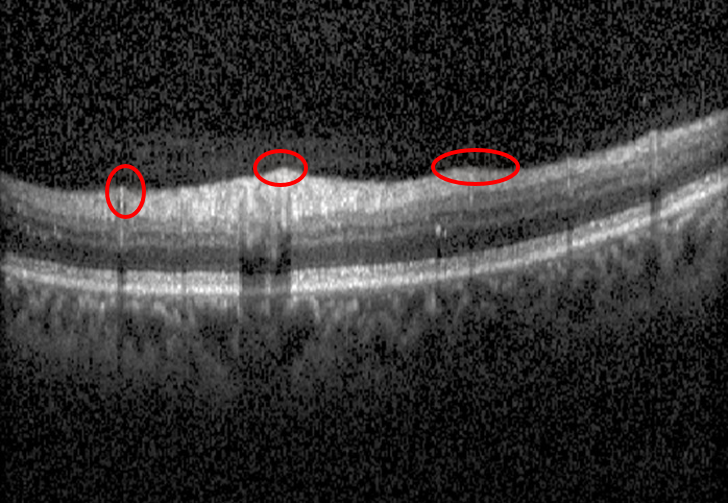}  \\
  \includegraphics[height=1.3cm, width=3.0cm]{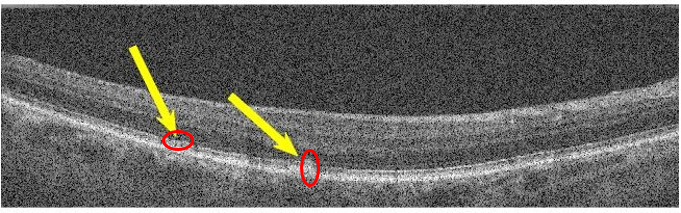} &
  \includegraphics[height=1.3cm, width=3.0cm]{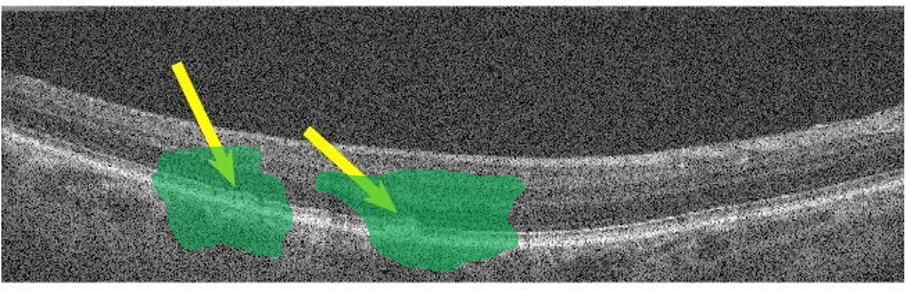} &
  \includegraphics[height=1.3cm, width=3.0cm]{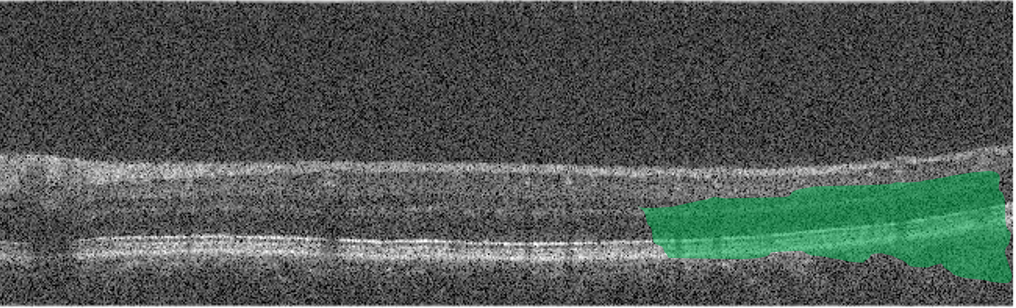} &
\includegraphics[height=1.3cm, width=3.0cm]{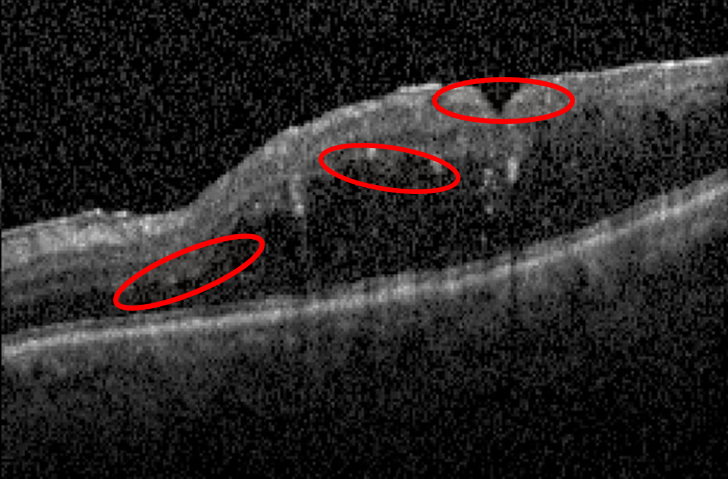} \\
\includegraphics[height=1.3cm, width=3.0cm]{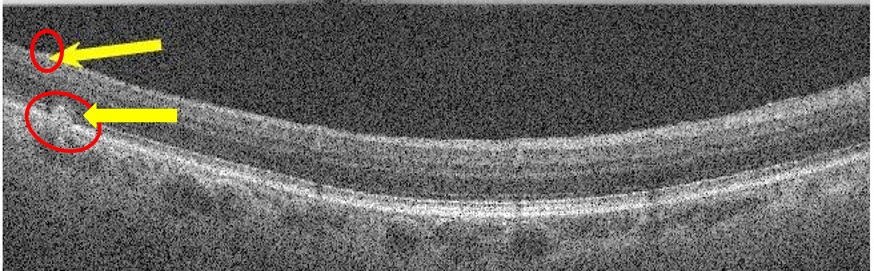} &
\includegraphics[height=1.3cm, width=3.0cm]{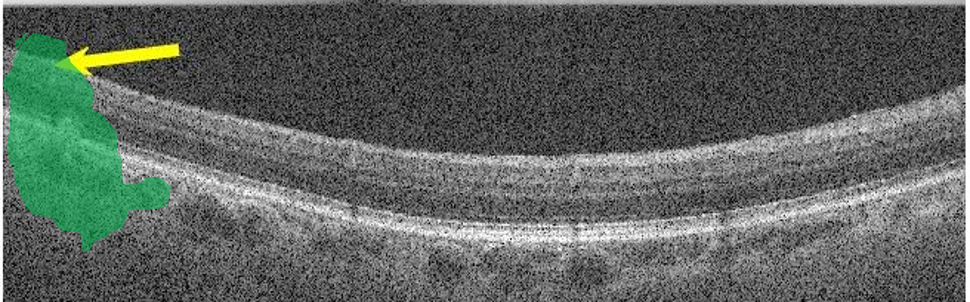} &
\includegraphics[height=1.3cm, width=3.0cm]{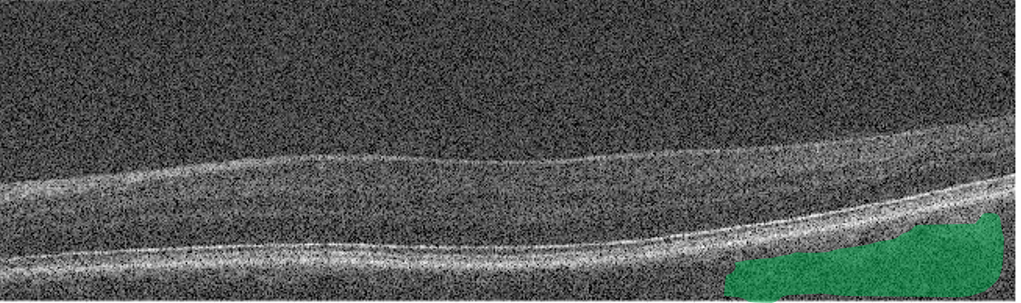} &
\includegraphics[height=1.3cm, width=3.0cm]{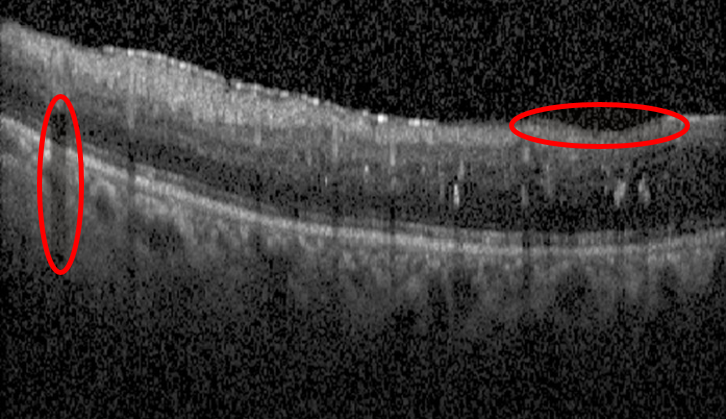}  \\
(a) & (b) & (c) & (d) \\
\end{tabular}
\caption{Example of misclassified images. Yellow arrows show positions of diseased activity in AMD images. (a) predictions by $Reg-DEC$ and quanitfication of disease activity; (b) CAMs by DenseNet; (c) normal images inaccurately classified as AMD by DenseNet with CAMs; (d) DME images correctly classified by $Reg-DEC$. Red circles are proportional to disease severity.}
\label{fig:ex}
\end{figure}

Figure~\ref{fig:ex} first and second columns, respectively, show AMD images accurately classified by $Reg-DEC$ and DenseNet. The yellow arrows highlight regions of abnormality identified by clinicians. Red ellipses (in first column) show the region of disease activity. The length of major axis quantifies magnitude of displacement of the corresponding grid point, and the orientation indicates direction. The local displacement magnitude is proportional to disease severity while the orientation identifies the exact location. The second column shows the corresponding CAMs obtained from DenseNet (region highlighted in green). Although the CAMs include the region of disease activity it does not localize it accurately and is spread out, nor does it output a measurable value. By dividing the displacement magnitude with the distance between the grid points we get a value very close to $p_d$. The advantages of our registration based method is obvious since it pinpoints abnormality and quantifies it in terms of displacement magnitude and angle.

Figure~\ref{fig:ex} third column shows examples of normal images that were rightly classified by $Reg-DEC$ but incorrectly classified as AMD by DenseNet. The green regions highlight disease activty as identified by DenseNet, which is erroneous since there are no abnormalities here. $Reg-DEC$ does not show any localization of pathologies in these examples. The fourth column shows examples of DME that were rightly identified by $Reg-DEC$, despite not being being part of the training data, alongwith red ellipses showing localized regions of disease activity.
They were assigned to clusters $4,6,7$ respectively. The CNNs trained to classify AMD and normal would mostly classify the second and third image as diseased while the first image was usually classified as normal  because of its similar appearance to some normal images. Thus, our method identifies different patient cohorts despite those not being part of the training data.

\section{Conclusion}
\label{sec:concl}

We propose a method to predict disease severity from retinal OCT images despite there being no labels provided for the disease severity. CNN regressor predicts registration parameters for a given test image which are undergo clustering to output a disease severity scale and a disease probability score in addition to the classification label (diseased or normal). Experimental results show our proposed method achieves better registration and classification performance compared to existing approaches. We are able to identify distinct patient cohorts not part of training data. Our approach also provides explainability behind the classification decision by quantifying disease activity from the registration parameters.

%

\bibliographystyle{splncs03}
\bibliography{MICCAI2019_OCTReg_Ref}

\begin{thebibliography}{10}
\providecommand{\url}[1]{\texttt{#1}}
\providecommand{\urlprefix}{URL }

\bibitem{Kaggle}
https://www.eyepacs.com

\bibitem{ITK}
"the insight segmentation and registration toolkit" www.itk.org

\bibitem{BalaCVPR18}
Balakrishnan, G., Zhao, A., Sabuncu, M., Guttag, J.: An supervised learning
  model for deformable medical image registration. In: Proc. CVPR. pp.
  9252--9260 (2018)

\bibitem{Mahapatra_GAN_CVIU2019}
Bozorgtabar, B., Mahapatra, D., von Teng, H., Pollinger, A., Ebner, L., Thiran,
  J.P., M.Reyes.: Informative sample generation using class aware generative
  adversarial networks for classification of chest xrays. Computer Vision and
  Image Understanding  184,  57--65 (2019)

\bibitem{Mahapatra_ISR_CMIG2019}
D.~Mahapatra, B.B., Garnavi, R.: Image super-resolution using progressive
  generative adversarial networks for medical image analysis. Computerized
  Medical Imaging and Graphics  71(1),  30--39 (2019)

\bibitem{FlowNet}
Dosovitskiy, A., Fischer, P., et. al.: Flownet: Learning optical flow with
  convolutional networks. In: In Proc. IEEE ICCV. pp. 2758--2766 (2015)

\bibitem{PathImimic18}
Graziani, M., Andrearczyk, V., Müller, H.: Regression concept vectors for
  bidirectional explanations in histopathology. In: In Proc. MICCAI-iMIMIC. pp.
  124--132 (2018)

\bibitem{ResNet}
He, K., Zhang, X., Ren, S., Sun, J.: Deep residual learning for image
  recognition. In: In Proc. CVPR (2016)

\bibitem{Densenet}
Huang, G., Liu, Z., van~der Maaten, L., Weinberger, K.: Densely connected
  convolutional networks. In: https://arxiv.org/abs/1608.06993, (2016)

\bibitem{KuangAMM14}
Kuang, H., Guthier, B., Saini, M., Mahapatra, D., Saddik, A.E.: A real-time
  smart assistant for video surveillance through handheld devices. In: In Proc:
  ACM Intl. Conf. Multimedia. pp. 917--920 (2014)

\bibitem{LiTMI_2015}
Li, Z., Mahapatra, D., J.Tielbeek, Stoker, J., van Vliet, L., Vos, F.: Image
  registration based on autocorrelation of local structure. IEEE Trans. Med.
  Imaging  35(1),  63--75 (2016)

\bibitem{MahapatraMiccaiIAHBD11}
Mahapatra, D.: Neonatal brain mri skull stripping using graph cuts and shape
  priors. In: In Proc: MICCAI workshop on Image Analysis of Human Brain
  Development (IAHBD) (2011)

\bibitem{MahapatraMLMI12}
Mahapatra, D.: Cardiac lv and rv segmentation using mutual context information.
  In: Proc. MICCAI-MLMI. pp. 201--209 (2012)

\bibitem{MahapatraGRSPIE12}
Mahapatra, D.: Groupwise registration of dynamic cardiac perfusion images using
  temporal information and segmentation information. In: In Proc: SPIE Medical
  Imaging (2012)

\bibitem{MahapatraSTACOM12}
Mahapatra, D.: Landmark detection in cardiac mri using learned local image
  statistics. In: Proc. MICCAI-Statistical Atlases and Computational Models of
  the Heart. Imaging and Modelling Challenges (STACOM). pp. 115--124 (2012)

\bibitem{MahapatraJDISkull2012}
Mahapatra, D.: Skull stripping of neonatal brain mri: Using prior shape
  information with graphcuts. J. Digit. Imaging  25(6),  802--814 (2012)

\bibitem{MahapatraJDIGCSP2013}
Mahapatra, D.: Cardiac image segmentation from cine cardiac mri using graph
  cuts and shape priors. J. Digit. Imaging  26(4),  721--730 (2013)

\bibitem{MahapatraJDIMutCont2013}
Mahapatra, D.: Cardiac mri segmentation using mutual context information from
  left and right ventricle. J. Digit. Imaging  26(5),  898--908 (2013)

\bibitem{MahapatraProISBI13}
Mahapatra, D.: Graph cut based automatic prostate segmentation using learned
  semantic information. In: Proc. IEEE ISBI. pp. 1304--1307 (2013)

\bibitem{MahapatraJDIJSGR2013}
Mahapatra, D.: Joint segmentation and groupwise registration of cardiac
  perfusion images using temporal information. J. Digit. Imaging  26(2),
  173--182 (2013)

\bibitem{MahapatraJDI_Cardiac_FSL}
Mahapatra, D.: Automatic cardiac segmentation using semantic information from
  random forests. J. Digit. Imaging.  27(6),  794--804 (2014)

\bibitem{Mahapatra_LME_CVIU}
Mahapatra, D.: Combining multiple expert annotations using semi-supervised
  learning and graph cuts for medical image segmentation. Computer Vision and
  Image Understanding  151(1),  114--123 (2016)

\bibitem{Mahapatra_LME_PR2017}
Mahapatra, D.: Semi-supervised learning and graph cuts for consensus based
  medical image segmentation. Pattern Recognition  63(1),  700--709 (2017)

\bibitem{Mahapatra_MICCAI17}
Mahapatra, D., Bozorgtabar, S., Hewavitahranage, S., Garnavi, R.: Image super
  resolution using generative adversarial networks and local saliencymaps for
  retinal image analysis,. In: In Proc. MICCAI. pp. 382--390 (2017)

\bibitem{MahapatraAL_MICCAI18}
Mahapatra, D., Bozorgtabar, S., Thiran, J.P., Reyes, M.: Efficient active
  learning for image classification and segmentation using a sample selection
  and conditional generative adversarial network. In: In Proc. MICCAI (2). pp.
  580--588 (2018)

\bibitem{Mahapatra_OMIA15}
Mahapatra, D., Buhmann, J.: Obtaining consensus annotations for retinal image
  segmentation using random forest and graph cuts. In: In Proc. OMIA. pp.
  41--48 (2015)

\bibitem{Mahapatra_MLMI15_Prostate}
Mahapatra, D., Buhmann, J.: Visual saliency based active learning for prostate
  mri segmentation. In: In Proc. MLMI. pp. 9--16 (2015)

\bibitem{Mahapatra_SSLAL_Pro_JMI}
Mahapatra, D., Buhmann, J.: Visual saliency based active learning for prostate
  mri segmentation. SPIE Journal of Medical Imaging  3(1) (2016)

\bibitem{MahapatraRVISBI13}
Mahapatra, D., Buhmann, J.: Automatic cardiac rv segmentation using semantic
  information with graph cuts. In: Proc. IEEE ISBI. pp. 1094--1097 (2013)

\bibitem{MahapatraTIP_RF2014}
Mahapatra, D., Buhmann, J.: Analyzing training information from random forests
  for improved image segmentation. IEEE Trans. Imag. Proc.  23(4),  1504--1512
  (2014)

\bibitem{MahapatraTBME_Pro2014}
Mahapatra, D., Buhmann, J.: Prostate mri segmentation using learned semantic
  knowledge and graph cuts. IEEE Trans. Biomed. Engg.  61(3),  756--764 (2014)

\bibitem{MahapatraISBI15_Optic}
Mahapatra, D., Buhmann, J.: A field of experts model for optic cup and disc
  segmentation from retinal fundus images. In: In Proc. IEEE ISBI. pp. 218--221
  (2015)

\bibitem{Mahapatra_ISBI19}
Mahapatra, D., Ge, Z.: Training data independent image registration with gans
  using transfer learning and segmentation information. In: In Proc. IEEE ISBI
  (2019)

\bibitem{Mahapatra_MLMI18}
Mahapatra, D., Ge, Z., Sedai, S., Chakravorty., R.: Joint registration and
  segmentation of xray images using generative adversarial networks. In: In
  Proc. MICCAI-MLMI. pp. 73--80 (2018)

\bibitem{Mahapatra_JSTSP2014}
Mahapatra, D., Gilani, S., Saini., M.: Coherency based spatio-temporal saliency
  detection for video object segmentation. IEEE Journal of Selected Topics in
  Signal Processing.  8(3),  454--462 (2014)

\bibitem{MahapatraTMI_CD2013}
Mahapatra, D., J.Tielbeek, Makanyanga, J., Stoker, J., Taylor, S., Vos, F.,
  Buhmann, J.: Automatic detection and segmentation of crohn's disease tissues
  from abdominal mri. IEEE Trans. Med. Imaging  32(12),  1232--1248 (2013)

\bibitem{MahapatraISBI_CD2014}
Mahapatra, D., J.Tielbeek, Makanyanga, J., Stoker, J., Taylor, S., Vos, F.,
  Buhmann, J.: Active learning based segmentation of crohn's disease using
  principles of visual saliency. In: Proc. IEEE ISBI. pp. 226--229 (2014)

\bibitem{Mahapatra_ABD2014}
Mahapatra, D., J.Tielbeek, Makanyanga, J., Stoker, J., Taylor, S., Vos, F.,
  Buhmann, J.: Combiningmultiple expert annotations using semi-supervised
  learning and graph cuts for crohn�s disease segmentation. In: In Proc:
  MICCAI-ABD (2014)

\bibitem{MahapatraJDICD2013}
Mahapatra, D., J.Tielbeek, Vos, F., Buhmann, J.: A supervised learning approach
  for crohn's disease detection using higher order image statistics and a novel
  shape asymmetry measure. J. Digit. Imaging  26(5),  920--931 (2013)

\bibitem{MahapatraISBI15_JSGR}
Mahapatra, D., Li, Z., Vos, F., Buhmann, J.: Joint segmentation and groupwise
  registration of cardiac dce mri using sparse data representations. In: In
  Proc. IEEE ISBI. pp. 1312--1315 (2015)

\bibitem{MahapatraICIT06}
Mahapatra, D., Routray, A., Mishra, C.: An active snake model for
  classification of extreme emotions. In: IEEE International Conference on
  Industrial Technology (ICIT). pp. 2195--2199 (2006)

\bibitem{Mahapatra_EMBC16}
Mahapatra, D., Roy, P., Sedai, S., Garnavi, R.: A cnn based neurobiology
  inspired approach for retinal image quality assessment. In: In Proc. EMBC.
  pp. 1304--1307 (2016)

\bibitem{Mahapatra_MLMI16}
Mahapatra, D., Roy, P., Sedai, S., Garnavi, R.: Retinal image quality
  classification using saliency maps and cnns. In: In Proc. MICCAI-MLMI. pp.
  172--179 (2016)

\bibitem{MahapatraICBME08_Retrieve}
Mahapatra, D., Roy, S., Sun, Y.: Retrieval of mr kidney images by incorporating
  spatial information in histogram of low level features. In: In 13th
  International Conference on Biomedical Engineering (2008)

\bibitem{MahapatraICME08}
Mahapatra, D., Saini, M., Sun, Y.: Illumination invariant tracking in office
  environments using neurobiology-saliency based particle filter. In: IEEE
  ICME. pp. 953--956 (2008)

\bibitem{MahapatraMICCAI_CD2013}
Mahapatra, D., Sch$\ddot{u}$ffler, P., Tielbeek, J., Vos, F., Buhmann, J.:
  Semi-supervised and active learning for automatic segmentation of crohn's
  disease. In: Proc. MICCAI, Part 2. pp. 214--221 (2013)

\bibitem{MahapatraMiccai08}
Mahapatra, D., Sun, Y.: Nonrigid registration of dynamic renal {MR} images
  using a saliency based {MRF} model. In: Proc. MICCAI. pp. 771--779 (2008)

\bibitem{MahapatraISBI08}
Mahapatra, D., Sun, Y.: Registration of dynamic renal {MR} images using
  neurobiological model of saliency. In: Proc. ISBI. pp. 1119--1122 (2008)

\bibitem{MahapatraICBME08_Sal}
Mahapatra, D., Sun, Y.: Using saliency features for graphcut segmentation of
  perfusion kidney images. In: In 13th International Conference on Biomedical
  Engineering (2008)

\bibitem{MahapatraMiccai10}
Mahapatra, D., Sun, Y.: Joint registration and segmentation of dynamic cardiac
  perfusion images using mrfs. In: Proc. MICCAI. pp. 493--501 (2010)

\bibitem{MahapatraICIP10}
Mahapatra, D., Sun., Y.: An mrf framework for joint registration and
  segmentation of natural and perfusion images. In: Proc. IEEE ICIP. pp.
  1709--1712 (2010)

\bibitem{MahapatraICDIP10a}
Mahapatra, D., Sun, Y.: Retrieval of perfusion images using cosegmentation and
  shape context information. In: Proc. APSIPA Annual Summit and Conference
  (ASC) (2010)

\bibitem{MahapatraEURASIP2010}
Mahapatra, D., Sun, Y.: Rigid registration of renal perfusion images using a
  neurobiology based visual saliency model. EURASIP Journal on Image and Video
  Processing. pp. 1--16 (2010)

\bibitem{MahapatraICDIP10b}
Mahapatra, D., Sun, Y.: A saliency based mrf method for the joint registration
  and segmentation of dynamic renal mr images. In: Proc. ICDIP (2010)

\bibitem{MahapatraTBME2011}
Mahapatra, D., Sun, Y.: Mrf based intensity invariant elastic registration of
  cardiac perfusion images using saliency information. IEEE Trans. Biomed.
  Engg.  58(4),  991--1000 (2011)

\bibitem{MahapatraMiccai11}
Mahapatra, D., Sun, Y.: Orientation histograms as shape priors for left
  ventricle segmentation using graph cuts. In: In Proc: MICCAI. pp. 420--427
  (2011)

\bibitem{MahapatraTIP2012}
Mahapatra, D., Sun, Y.: Integrating segmentation information for improved
  mrf-based elastic image registration. IEEE Trans. Imag. Proc.  21(1),
  170--183 (2012)

\bibitem{MahapatraABD12}
Mahapatra, D., Tielbeek, J., Buhmann, J., Vos, F.: A supervised learning based
  approach to detect crohn's disease in abdominal mr volumes. In: Proc. MICCAI
  workshop Computational and Clinical Applications in Abdominal
  Imaging(MICCAI-ABD). pp. 97--106 (2012)

\bibitem{MahapatraCDFssISBI13}
Mahapatra, D., Tielbeek, J., Vos, F., ., J.B.: Crohn's disease tissue
  segmentation from abdominal mri using semantic information and graph cuts.
  In: Proc. IEEE ISBI. pp. 358--361 (2013)

\bibitem{MahapatraCDSPIE13}
Mahapatra, D., Tielbeek, J., Vos, F., Buhmann, J.: Localizing and segmenting
  crohn's disease affected regions in abdominal mri using novel context
  features. In: Proc. SPIE Medical Imaging (2013)

\bibitem{MahapatraWssISBI13}
Mahapatra, D., Tielbeek, J., Vos, F., Buhmann, J.: Weakly supervised semantic
  segmentation of crohn's disease tissues from abdominal mri. In: Proc. IEEE
  ISBI. pp. 832--835 (2013)

\bibitem{MahapatraISBI15_CD}
Mahapatra, D., Vos, F., Buhmann, J.: Crohn's disease segmentation from mri
  using learned image priors. In: In Proc. IEEE ISBI. pp. 625--628 (2015)

\bibitem{Mahapatra_SSLAL_CD_CMPB}
Mahapatra, D., Vos, F., Buhmann, J.: Active learning based segmentation of
  crohns disease from abdominal mri. Computer Methods and Programs in
  Biomedicine  128(1),  75--85 (2016)

\bibitem{MahapatraSPIE08}
Mahapatra, D., Winkler, S., Yen, S.: Motion saliency outweighs other low-level
  features while watching videos. In: SPIE HVEI. pp. 1--10 (2008)

\bibitem{BrainImimic18}
Pereira, S., Meier, R., Alves, V., Reyes, M., Silva., C.: Automatic brain tumor
  grading from mri data using convolutional neural networks and quality
  assessment. In: In Proc. MICCAI-iMIMIC. pp. 106--114 (2018)

\bibitem{AMDData}
Rasti, R., Rabbani, H., Mehri, A., Hajizadeh, F.: Macular oct classification
  using a multi-scale convolutional neural network ensemble. IEEE Trans. Med.
  Imag.  37(4),  1024--1034 (2018)

\bibitem{VGG}
Simonyan, K., Zisserman., A.: Very deep convolutional networks for large-scale
  image recognition. CoRR  abs/1409.1556 (2014)

\bibitem{RegNet}
Sokooti, H., de~Vos, B., Berendsen, F., Lelieveldt, B., Isgum, I., Staring, M.:
  Nonrigid image registration using multiscale 3d convolutional neural
  networks. In: MICCAI. pp. 232--239 (2017)

\bibitem{Vincent2010}
Vincent, P., Larochelle, H., Lajoie, I., Bengio, Y., Manzagol, P.: Stacked
  denoising autoencoders: Learning useful representations in a deep network
  with a local denoising criterion. Journal of Mach. Learn. Res.  11,
  3371--3408 (2010)

\bibitem{Vos_DIR}
de~Vos, B., Berendsen, F., Viergever, M., Staring, M., Isgum, I.: End-to-end
  unsupervised deformable image registration with a convolutional neural
  network. In: arXiv preprint arXiv:1704.06065 (2017)

\bibitem{VosEMBC}
Vos, F.M., Tielbeek, J., Naziroglu, R., Li, Z., Sch$\ddot{u}$ffler, P.,
  Mahapatra, D., Wiebel, A., Lavini, C., Buhmann, J., Hege, H., Stoker, J., van
  Vliet, L.: Computational modeling for assessment of {IBD}: to be or not to
  be? In: Proc. IEEE EMBC. pp. 3974--3977 (2012)

\bibitem{DEC}
Xie, J., Girshick, R., Farhadi, A.: Unsupervised deep embedding for clustering
  analysis. In: Proc. ICML. pp. 478--487 (2016)

\bibitem{CAM}
Zhou, B., Khosla, A., Lapedriza, A., Oliva, A., Torralba, A.: Learning deep
  features for discriminative localization. In: Proc. CVPR. pp. 2921--2929
  (2016)

\bibitem{Mahapatra_MLMI15_Optic}
Zilly, J., Buhmann, J., Mahapatra, D.: Boosting convolutional filters with
  entropy sampling for optic cup and disc image segmentation from fundus
  images. In: In Proc. MLMI. pp. 136--143 (2015)

\bibitem{Zilly_CMIG_2016}
Zilly, J., Buhmann, J., Mahapatra, D.: Glaucoma detection using entropy
  sampling and ensemble learning for automatic optic cup and disc segmentation.
  In Press Computerized Medical Imaging and Graphics  55(1),  28--41 (2017)

\end{thebibliography}

\end{document}